\documentclass[11pt]{article}

\usepackage[margin=1in]{geometry}      
\usepackage[utf8]{inputenc}
\usepackage[T1]{fontenc}               
\usepackage{amsmath, amssymb}          
\usepackage{graphicx}                  
\usepackage{url}                       
\usepackage{booktabs}
\usepackage{multirow}
\usepackage{adjustbox}
\usepackage{array}
\usepackage{makecell}
\newcolumntype{C}[1]{>{\centering\arraybackslash}p{#1}}
\usepackage{xcolor}
\definecolor{stdgray}{gray}{0.4}   
\newcommand{\std}[1]{\textcolor{stdgray}{$\pm$ #1}}

\graphicspath{{./}}   

\begin{document}

\begin{center}
{\Large\bfseries ImageCAS-X: a dataset and benchmark for coronary artery segmentation and centerline extraction in coronary CT angiography}
\end{center}

\bigskip

\begin{center}
Kit M. Bransby$^{1, \ast}$,
Esther {\O}ksnebjerg$^{1}$,
Kristoffer Kj{\ae}r$^{1}$,
Jacob Kirkeby$^{1}$,
Yasmin El Youssef$^{2}$,
A{\"i}da Jim{\'e}nez$^{1}$,
Philip R. Pedersson$^{2}$,
Martina C. de Knegt$^{2}$,
Klaus F. Kofoed$^{2}$,
Rasmus R. Paulsen$^{1, \ast}$
 
\medskip
\small
$^{1}$DTU Compute, Technical University of Denmark, Kongens Lyngby, Denmark \\
$^{2}$Cardiovascular Research Unit, Copenhagen University Hospital -- Rigshospitalet, Copenhagen, Denmark
 
\medskip
$^{\ast}$Corresponding authors: Kit M. Bransby \texttt{kimbr@dtu.dk} \& Rasmus R. Paulsen \texttt{rapa@dtu.dk}
\end{center}

\bigskip

\section*{Graphical Abstract}

\begin{figure}[h]
    \centering
    \includegraphics[width=1\linewidth]{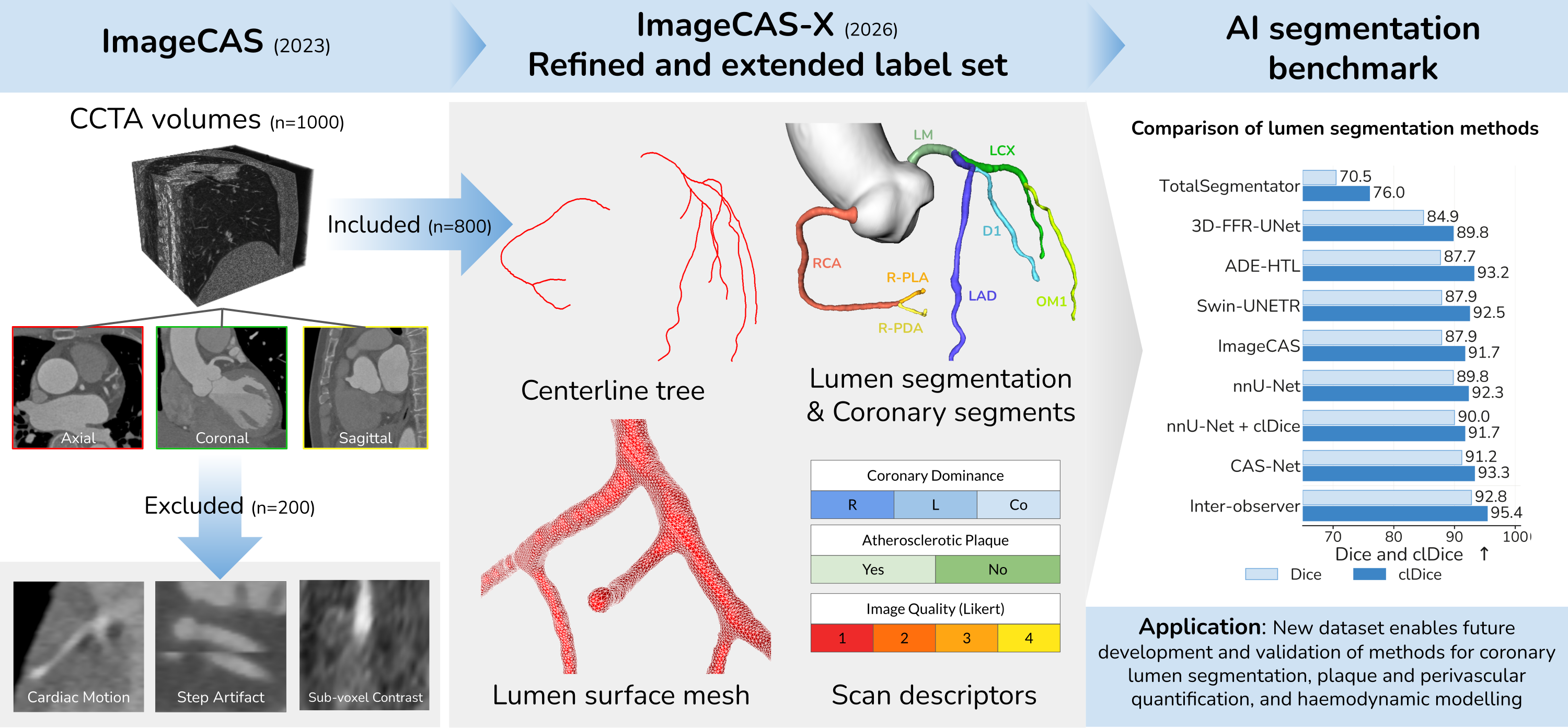}
\end{figure}

\section*{Abstract}

Accurate segmentation of the coronary vessel lumen is a prerequisite for quantitative assessment of atherosclerotic plaque and perivascular adipose tissue in coronary computed tomography angiography (CCTA). Cardiologists rely on semi-automated methods for this task because manual vessel tracing and segmentation are labour-intensive. Although many automated methods have been proposed, their validation remains limited by the lack of large, high-quality publicly available datasets. We provide a new dataset of voxel-wise annotations of the vessel lumen and coronary segments, alongside centerlines, and mesh surfaces for 800 scans from the publicly available ImageCAS dataset. Using this dataset, we benchmark established lumen segmentation methods against inter-observer variability, stratifying performance by disease, image quality, coronary dominance, coronary segment, vessel diameter, and lumen attenuation. These labels allow segmentation accuracy to be described in anatomical and clinical context rather than reported as a single aggregate score. The dataset supports the development and validation of methods for lumen segmentation, plaque and perivascular quantification, and haemodynamic modelling.

\bigskip

\section*{Background \& Summary}

Coronary computed tomography angiography (CCTA) is the primary non-invasive imaging modality for the assessment of coronary artery disease (CAD). It enables three-dimensional visualisation of the coronary arteries and differentiation of the lumen, atherosclerotic plaque components, and surrounding perivascular adipose tissue~\cite{leipsic_scct_2014,nieman_standards_2024}. CCTA is recommended as a first-line diagnostic test for suspected CAD and has an emerging role in quantitative plaque assessment, longitudinal monitoring, and identifying haemodynamically significant lesions~\cite{knuuti20202019,writing20212021}. Beyond routine clinical use, CCTA has also driven advances in CAD research through imaging-derived biomarkers such as high-risk plaque features~\cite{scotheart, williams_coronary_2019,thomsen2016characteristics}, fat attenuation index for perivascular inflammation~\cite{oikonomou_non-invasive_2018,antonopoulos_detecting_2017}, and computational haemodynamic modelling~\cite{norgaard_diagnostic_2014}, which are often used in clinical trials and population studies.

These applications rely on anatomically precise delineation of the coronary lumen; however, achieving such contours in CCTA remains technically challenging due to partial volume effects, motion artefacts, blooming artefacts from calcification, and limited spatial resolution~\cite{leipsic_scct_2014,nieman_standards_2024}. In addition, segmentation algorithms must preserve the continuous, branching topology of the coronary tree, as even small discontinuities can compromise downstream quantitative analysis. Expert annotation is the clinical reference standard, but it is labour-intensive and subject to inter-observer variability, limiting scalability and reproducibility. Despite methodological advances in vessel segmentation algorithms~\cite{dong2023novel,song2022automatic,qiu2025topology,yang2024segmentation,shit2021cldice}, current guidelines continue to recommend semi-automated workflows, reflecting the absence of sufficiently validated fully automated solutions~\cite{acc,nieman_standards_2024}. Progress towards this end requires large, high-quality, publicly available datasets that allow for rigorous evaluation, benchmarking, and comparison across methods. 

Several CCTA datasets have supported the development of coronary tracing and segmentation methods~\cite{metz20083d,kiricsli2013standardized,gharleghi_annotated_2023,tu_mask_2025,zeng_imagecas_2023,yang2024segmentation}. Early public benchmarks, such as the MICCAI 2008 coronary artery tracking challenge (CAT08)~\cite{metz20083d} and the MICCAI 2012 cardiovascular imaging challenge~\cite{kiricsli2013standardized}, established standardised evaluation protocols for coronary centerline tracing and lumen segmentation. However, with only 32 and 48 cases respectively, these datasets were limited in scale and are no longer publicly available. More recently, the ASOCA dataset~\cite{gharleghi_annotated_2023} released a high-quality label set of lumen segmentations for healthy and diseased patients, but still comprises only 60 cases. Larger cohorts have since been curated. PCCTA120~\cite{tu_mask_2025} provided 120 scans with inner and outer wall annotations, yet has not been publicly released. ImageCAS~\cite{zeng_imagecas_2023}, currently the largest dataset with 1,000 scans, released voxel-wise lumen masks but lacks sufficient segmentation accuracy for reliable benchmarking and downstream applications (examples provided in Supplementary section 1). To date, no dataset combines large-scale CCTA imaging with high-quality lumen annotations suitable for training and validating vessel segmentation methods.

To address this gap, we provide a large-scale label set for the publicly available ImageCAS dataset, comprising high-quality voxel-wise segmentations of the coronary lumen, alongside mesh surfaces and centerlines. To demonstrate its utility, the label set is used to train, validate, and benchmark state-of-the-art vessel segmentation algorithms, providing a systematic evaluation of their strengths and limitations. Current guidelines recommend that pathological findings be reported per vessel segment~\cite{leipsic_scct_2014}, reflecting that arteries differ in clinical importance according to the myocardial territory they supply. To support this, we provide additional labels for anatomical segment, disease, and coronary dominance, allowing performance to be assessed in clinically meaningful regions. 

This dataset enables several applications previously constrained by the lack of large-scale, high-quality lumen annotations. Whereas validation of segmentation pipelines has largely relied on proprietary, industry-held data, the public availability and balanced representation of diseased and non-diseased cases enable independent benchmarking for the first time at this scale. Accurate lumen contours are also a prerequisite for quantifying plaque burden, plaque composition, and perivascular tissue attenuation; because earlier datasets were neither large nor precise enough to validate contouring, our generated data now support the development of plaque and perivascular segmentation algorithms. The dataset also enables validation of automated methods against human performance: by benchmarking algorithms against inter-observer variability, methods can be shown to operate within the range of human agreement which is a prerequisite for deployment in clinical trials and population studies. The accompanying mesh surfaces further enable computational modelling of the coronaries including computed tomography fractional flow reserve (CT-FFR), haemodynamic simulation, and stent planning algorithms that depend on accurate vessel geometry. Beyond coronary-specific analysis, we anticipate the dataset will be of broader value to the medical imaging and machine learning communities, supporting research in topology-aware segmentation~\cite{shit2021cldice,clough2020} and synthetic data generation~\cite{bransby2023,wang2025deepca,feldman2025}.

\section*{Methods}

\subsection*{Patient Cohort \& Imaging}

We retrospectively analysed CCTA data from 1,000 patients collected for the open-source ImageCAS dataset~\cite{zeng_imagecas_2023} at Guangdong Provincial People's Hospital between 2012 and 2018. The patients had a documented medical history of ischaemic stroke, transient ischaemic attack, or peripheral artery disease, and included some patients who underwent early revascularisation (within 90 days). The cohort contains both males (n = 586) and females (n = 414), of a minimum age of 18 and average ages of 57.7 and 60.0 years, respectively. All patients underwent CCTA acquired using a Siemens 128-slice dual-source scanner, and either the 30--40\% or 60--70\% cardiac phase was selected based on optimal coronary image quality. The resulting image volumes contained $512 \times 512 \times (206\text{–}275)$ voxels, with an in-plane pixel spacing of 0.29--0.43 mm and slice spacing of 0.25--0.45 mm. The original collection of CCTA data received Research Ethics Committee (REC) approval from Guangdong Provincial People’s Hospital, Guangdong Academy of Medical Sciences under Protocol No. 2019324H. It complied with all relevant ethical regulations and the data was anonymised so that sensitive information including name, date of birth, and hospital meta-data were all removed. Our secondary use of this data did not collect any new human-subjects data, and as the data was already anonymised, no ethical approval was required. 

\subsection*{Annotation Protocol}

\begin{figure}[h]
    \centering
    \includegraphics[width=1\linewidth]{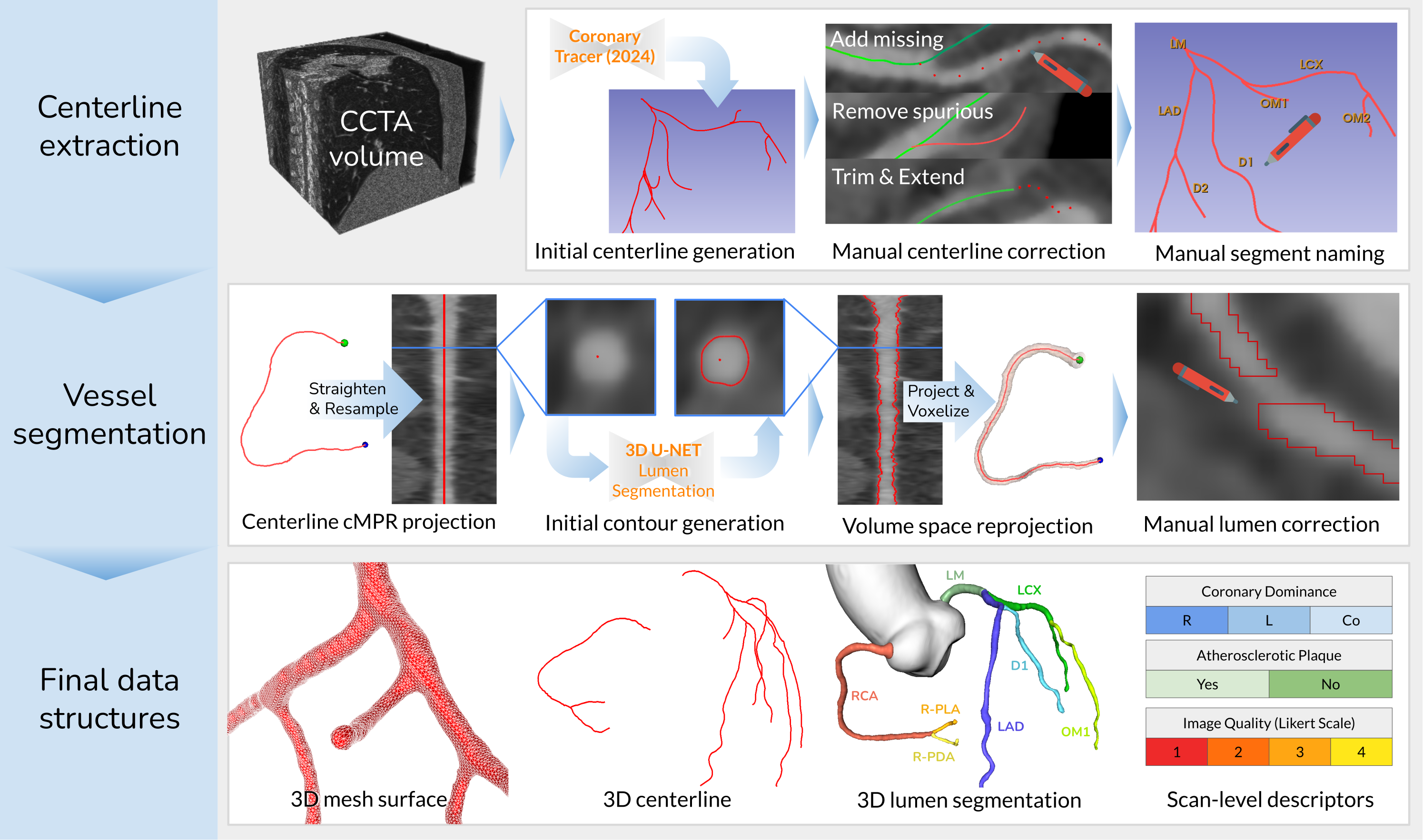}
    \caption{ImageCAS-X data creation pipeline. Top: initial centerlines are generated automatically and manually corrected and named; Middle: the corrected centerlines are used to generate initial lumen contours in cMPR space which are voxelised and manually corrected; Bottom: 3D mesh surface, coronary centerline, lumen segmentations split by coronary segment name, and scan-level descriptors are provided for each patient.}
    \label{fig:data_creation}
\end{figure}

\subsubsection*{Software} 
CCTA analysis was performed using CoronaryExplorer (v0.1), a semi-automatic tool for labelling coronary vessel centerlines and vessel lumen segmentation, developed as a 3D Slicer (v5.10) extension for the purpose of analysing this dataset. The labelling protocol for ImageCAS using CoronaryExplorer is summarised in Fig.~\ref{fig:data_creation} and detailed below. 

\subsubsection*{Image Quality} 
An expert analyst reviewed all CCTA images and graded each on a five-point Likert scale: non-diagnostic (0), poor (1), adequate (2), good (3), and excellent (4). Images graded non-diagnostic, in which artefacts caused anatomically implausible or ambiguous vessel boundaries, were excluded. In total, 200 scans were excluded owing to motion artefacts (n = 114), step artefacts (n = 76), poor contrast mixing (n = 7), static noise (n = 1), incorrect field of view (n = 1), and file corruption (n = 1). Of the 800 scans retained, 55 were graded poor, 140 adequate, 168 good, and 437 excellent.

\subsubsection*{Coronary Dominance}
Coronary dominance of each scan was determined where right coronary dominance is defined by the presence of posterior descending artery (PDA) and posterior lateral artery (PLA) from the right coronary artery (RCA), and left dominance is defined by a PDA from the left circumflex artery (LCx). We define co-dominance by the presence of PDAs or PLAs from both RCA and LCx, or the presence of a PDA from the RCA without a PLA from the RCA. The 800 included scans consisted of 729 (91.1\%) right dominant, 41 (5.1\%) left dominant, and 30 (3.8\%) co-dominant cases. 

\subsubsection*{Coronary Artery Disease}

Each scan was assigned a binary label by the analyst to describe if the patient had coronary artery disease. This was achieved by identifying any atherosclerotic lesions composed of calcified, non-calcified or mixed plaque. Of the 800 included scans, 388 (48.5\%) were labelled as diseased, and 412 (51.5\%) as no disease. 

\subsubsection*{Centerline Extraction} 
The selected cases were divided equally among four trained analysts, who used the CoronaryExplorer tool to extract vessel centerlines and annotate the lumen. Generating coronary vessel annotations from scratch is time consuming and impractical at scale, therefore we follow standard workflows~\cite{nieman_standards_2024} by tracing centerlines using a previously validated automated  method~\cite{baldachowski_coronary_3d_ct}. The resulting centerlines contained errors and were then manually refined by trimming, removing spurious segments, and drawing any missing vessels, ensuring the final coronary tree conformed to the established 18-segment model~\cite{leipsic_scct_2014}. We do not distinguish between proximal, mid and distal segments of the left anterior descending (LAD) and left circumflex (LCx) because they depend on side branch positions which are highly variable, and for consistency we also consider the RCA as a single segment. An additional `Other' category was included for third and forth diagonal or obtuse marginal branches outside of the 18-segment model, which were > 1.8mm in diameter at the most proximal end or as large as the first or second branches. Finally, each branch segment was manually classified according to the 18-segment naming convention. In the Supplementary section 2, we provide further details on the protocol used to discretise variations in coronary anatomy. The total time taken for manual centerline correction was 200 hours. 

\subsubsection*{Vessel Segmentation} 
Each extracted centerline was used to generate a curved multiplanar reformatting (cMPR) volume by reparameterising the image volume in a coordinate system aligned with the vessel’s central axis and resampling image intensities onto planes orthogonal to the centerline, producing a straightened representation of the vessel~\cite{cpr}. The in-plane resolution was 0.25 mm with a 16 × 16 mm field of view, and cross-sections were sampled at 0.4 mm intervals along the vessel. The lumen in 100 randomly selected cMPR vessel volumes were manually annotated and used to train and validate a 3D U-Net (further details are provided in Supplementary section 3). The trained network was then applied to all scans to generate initial 3D lumen surface predictions in cMPR volume space. The lumen surface was projected from cMPR space back to the original CCTA volume space and voxelised resulting in a 3D lumen mask. The analysts subsequently reviewed and manually corrected the mask by inspecting each axial slice sequentially. All cases were reviewed by the lead analyst and any errors were corrected. The total time taken for lumen segmentation correction was 270 hours.

\begin{figure}[h]
    \centering
    \includegraphics[width=0.8\linewidth]{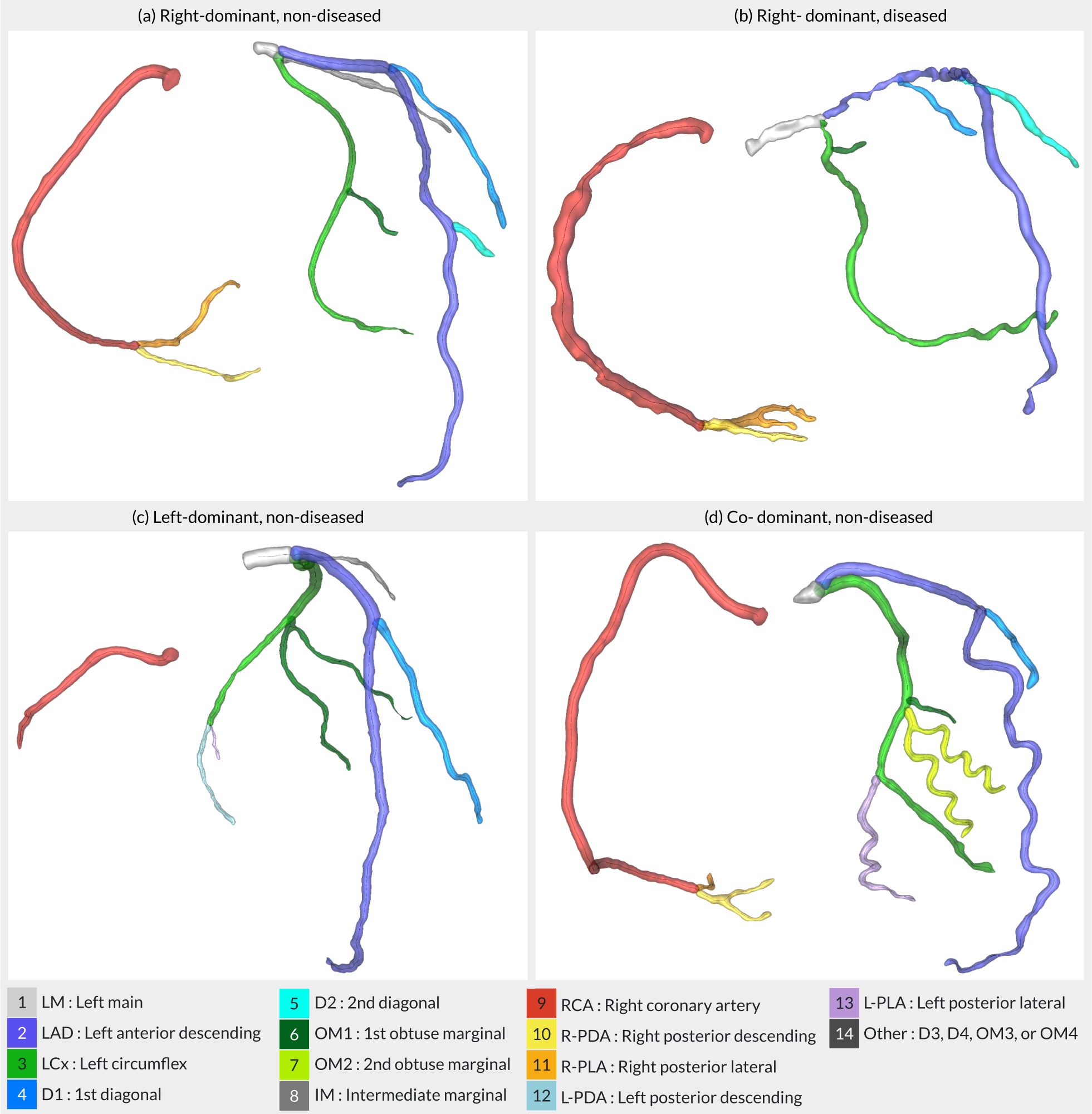}
    \caption{Sample coronary segments and centerlines for right-, left- and co-dominant trees and diseased versus non-diseased patients.}
    \label{fig:examples}
\end{figure}

\subsubsection*{Postprocessing}

Segment-labelled centerlines did not always pass through the centre of the corrected lumen contours, so new centerlines were generated from the corrected masks using a skeletonisation method~\cite{skeletonisation}, followed by Gaussian smoothing ($\sigma = 0.5$ mm) with a sliding window of 5 vertices. The aorta was segmented using TotalSegmentator's high resolution heart chambers model~\cite{totalsegmentator}, and start points were assigned to centerline vertices with a degree of 1 lying within 5 mm of the aorta. End points were assigned to all remaining vertices of degree 1, and bifurcation points to vertices with a degree of 3 or more. To recover the segment names, the new centerline tree was partitioned at bifurcation points into segments, which were then matched to the original segment-named centerlines by shortest distance. Segment assignments were manually reviewed by the lead analyst and any errors corrected. Segment names were propagated to every lumen voxel in the segmentation mask by assigning each voxel the name of its nearest centerline point. Mesh surfaces were generated from the corrected masks using the marching cubes algorithm in the Visualization Toolkit (VTK), followed by Taubin smoothing (50 iterations, passband = 0.05). Samples from the dataset are presented in Fig.~\ref{fig:examples}.

\subsubsection*{Data Subsets}

A total of 800 annotated cases were randomly split into training (70\%, 560 cases), validation (10\%, 80 cases), and test (20\%, 160 cases) sets. To assess inter-observer variability, each test case was additionally re-annotated by a different analyst, selected at random, following the same protocol without review from the lead analyst and blinded to the first set of labels.

\section*{Data Records}

The ImageCAS-X data generated for this study are publicly accessible via Zenodo (\url{https://zenodo.org/records/21887809})~\cite{bransby_zenodo}. The repository is organised by data type into subdirectories for segmentation masks, centerlines, mesh surfaces and file lists. Segmentation masks are stored as compressed NIfTI files in .nii.gz format, and centerlines and mesh surfaces as VTK files in .vtk format. Each centerline carries binary feature attributes that assign start, bifurcation, and end points to every vertex, and a categorical attribute that assigns a coronary segment to each vertex. Coronary segments are also encoded at the voxel level in the segmentation masks, where each voxel is assigned to a discrete class; the mapping from class index to coronary segment name is given in Supplementary section 4. The original ImageCAS volumes~\cite{zeng_imagecas_2023} are not re-distributed in our repository but are publicly available here: ~\url{https://www.kaggle.com/datasets/xiaoweixumedicalai/imagecas}. 

\begin{figure}
    \centering
    \includegraphics[width=1\linewidth]{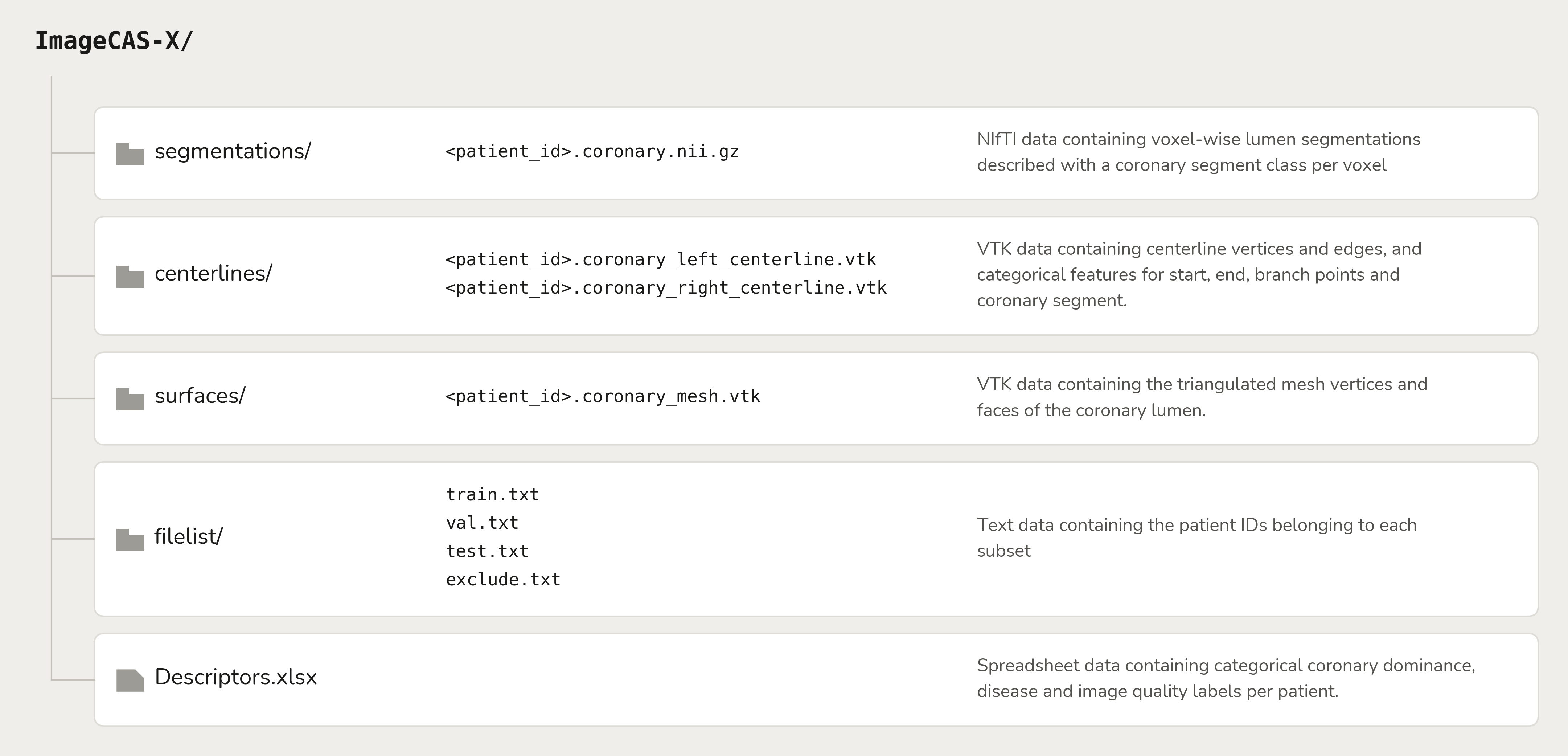}
    \caption{Summary of data structures in the ImageCAS-X repository}
    \label{fig:data_records}
\end{figure}

Each patient is assigned a unique ID, identical to the original ImageCAS 2023 dataset~\cite{zeng_imagecas_2023}, which prefixes every data file belonging to that patient. Subset membership is indicated by four text files (train, validation, test, exclude) in .txt format. Categorical labels for coronary dominance, disease, and image quality are provided in Descriptors.xlsx. A summary of this repository structure is presented in Fig.~\ref{fig:data_records}. 

\section*{Technical Validation}

\subsection*{Benchmark}

To demonstrate the practical utility and clinical relevance of the data, we trained and benchmarked state-of-the-art lumen segmentation methods. Works from high-impact journals published in the past six years were selected that represent the latest methodological contributions for lumen segmentation in CCTA. These works span three distinct categories:  patch-based segmentation (3D-FFR-UNet~\cite{song2022automatic}, CAS-Net~\cite{dong2023novel}) where patches are randomly sampled from the full volume, coarse-to-fine segmentation (ImageCAS~\cite{zeng_imagecas_2023}) where an initial dilated segmentation is used to focus patch segmentation on vessel areas only, and centerline-based segmentation (ADE-HTL~\cite{zhang2023anatomy}) where centerline bifurcation points and connectivity is used to guide lumen segmentation. In addition, we included nnU-Net~\cite{isensee2021nnu} and Swin-UNETR~\cite{swinunetr} as general-purpose medical image segmentation methods, nnU-Net with clDice~\cite{shit2021cldice} loss as a topology-focused method, and TotalSegmentator~\cite{totalsegmentator} as a zero-shot pretrained model. For all methods, we estimated the luminal centerline by skeletonising the voxel-based segmentation predictions. Three methods do not provide publicly available code~\cite{zeng_imagecas_2023,zhang2023anatomy,song2022automatic}; therefore we reimplemented each based on the written methodology. We excluded CorSegRec~\cite{qiu2025topology} and CCA~\cite{yang2024segmentation} as a reliable implementation was not possible from their written methodology, and DeformCL~\cite{deformCL} because it can only handle single branches. Implementation details of all methods are provided in Supplementary section 6. 

All experiments were implemented in PyTorch 2.9 and performed on an NVIDIA RTX 5090 GPU with 32 GB of memory. We standardised training and inference by adopting an nnU-Net style pipeline, where common hyperparameters such as preprocessing, resampling resolution, class sampling, augmentation, learning rate, optimiser, postprocessing, and test-time augmentation were fixed. Volumes were isotropically resampled to 0.5 mm and geometric and intensity-based augmentations were applied during training. We defined an epoch as 250 training mini-batches and 50 validation mini-batches, where batch size was maximised to fit the GPU memory, and the total number of epochs was 1,000. An Adam optimiser and polynomial learning rate scheduler were used for all experiments, apart from nnU-Net which used an SGD optimiser. All method-specific hyperparameters such as loss, input size and model configurations were set based on their original implementation. For each method, we tuned the initial learning rate using 0.01, 0.001 and 0.0001, and selected the model with the best validation DSC. The full training and evaluation recipe is provided in Supplementary section 5.   

To provide clinical context to the benchmark results and demonstrate the quality of segmentations, we also compared model performance to the inter-observer variability computed from the manually relabelled test set, and to the original ImageCAS labels~\cite{zeng_imagecas_2023}. Our benchmark and code are available at \url{https://kitbransby.github.io/ImageCAS-X/}. 

\subsection*{Statistical Measures}

All quantitative results are presented as mean $\pm$ standard deviation. To compare segmentations, the Dice similarity coefficient (DSC) was used to measure voxel overlap and 95th percentile Hausdorff distance (HD95) to measure large surface deviations~\cite{taha2015}, and the Betti number error ($\beta^{err}$) computed volumetrically, which compares topology between predicted and ground truth segmentations in two dimensions: the first counts the number of connected components and the second counts the number of loops~\cite{wasserman2018topological}. Centerline quality was evaluated using clDice~\cite{shit2021cldice} to measure topological error of skeletonised segmentations, average symmetric surface distance (ASSD) to describe the overall alignment and HD95 to measure large deviations, and both distances are reported in mm~\cite{taha2015}. 

To measure the effect of lumen attenuation, distal position and vessel diameter on segmentation performance, we sampled local patches centred at different positions on the ground truth coronary tree and computed a local DSC for each patch. Specifically, we used every fifth vertex (2 mm) of the ground truth centerline tree to build a $8  \times 8 \times 8$ mm cube centred on that vertex, and the local DSC is computed between the segmentation prediction and ground truth for that patch. The geodesic distance from the ostium, the lumen diameter, the mean lumen Hounsfield unit, and the coronary segment label of each patch was recorded, and patch metrics were binned for each variable within each patient and averaged, to show how the variable affected segmentation performance. 

All statistical tests were non-parametric and two-sided, with the patient as the unit of analysis. Method-vs-analyst differences on the same scans used the Wilcoxon signed-rank test; differences between independent groups of scans used the Mann--Whitney U test for two groups and the Kruskal--Wallis test for more. Comparison across coronary segments and across ordinal bins were repeated measures since each patient contributed a value to multiple bins, therefore these were assessed with the Skillings--Mack test. Monotonic trends across ordinal groups were described with Spearman's rank correlation ($\rho$). Significance was set at $p < 0.05$.

\subsection*{Quantitative Results}

\subsubsection*{Inter-observer variability}

For the inter-observer variability, the DSC was 92.8, HD95 was 2.46 mm and $\beta^{err}$ was 0.2 for lumen segmentation, while the clDice was 95.4, ASSD was 0.53 mm and the HD95 was 4.58 mm for centerline extraction (Table~\ref{tab:inter_segment}). When comparing our labels to the human annotators from the original ImageCAS dataset~\cite{zeng_imagecas_2023} who also segmented the lumen according the 18-segment model, DSC was 41.8, HD95 was 16.15 mm, and $\beta^{err}$ was 7.0 for lumen segmentation, while the clDice was 78.2, ASSD was 2.23 mm and the HD95 was 18.89 mm for centerline extraction (Table~\ref{tab:quant_main}). The differences to our manual annotations are significant ($p < 0.001$) and when coupled with qualitative examples (Supplementary Fig. 1) reflect the large improvement in lumen segmentation quality in the new dataset. 

\begin{table*}[h]
    \centering
    \begin{adjustbox}{width=0.9\linewidth}
    \begin{tabular}{lcccccccc}
    \toprule
    & & & \multicolumn{3}{c}{Lumen Segmentation} & \multicolumn{3}{c}{Lumen Centerline} \\
    \cmidrule(lr){4-6} \cmidrule(lr){7-9}
    Segment & $n$ & Agreement$\uparrow$ & DSC$\uparrow$ & HD95$\downarrow$ &
  $\beta^{err}$$\downarrow$ &
    clDice$\uparrow$ & ASSD$\downarrow$ & HD95$\downarrow$  \\
    \midrule
LM &
   155 & 96.9 & 91.9 \std{13.7} & 1.41 \std{5.58} & 0.00 \std{0.00} & 95.6\std{17.6} & 0.40
  \std{1.52} & 1.45 \std{5.79} \\
  LAD &
   160 & 100.0 & 92.3 \std{6.7} & 3.39 \std{9.07} & 0.10 \std{0.36} & 96.4 \std{7.9} & 0.48
  \std{1.29} & 3.90 \std{9.67} \\
  LCx &
   159 & 99.4 & 84.8 \std{19.8} & 8.59 \std{15.30} & 0.08 \std{0.29} & 87.3 \std{23.0} & 1.85
  \std{3.86} & 9.02 \std{15.80} \\
  D1 &
   155 & 96.9 & 79.9 \std{28.7} & 6.66 \std{12.30} & 0.04 \std{0.19} & 85.0 \std{30.2} & 2.53
  \std{6.80} & 6.83 \std{12.40} \\
  D2 &
   91 & 89.4 & 82.9 \std{24.3} & 3.96 \std{6.92} & 0.02 \std{0.15} & 88.1 \std{25.3} & 1.36
  \std{3.97} & 4.10 \std{7.01} \\
  OM1 &
   132 & 94.4 & 74.1 \std{32.3} & 8.93 \std{15.70} & 0.05 \std{0.21} & 79.8 \std{34.2} & 3.47
  \std{8.27} & 9.21 \std{16.00} \\
  OM2 &
   46 & 87.5 & 77.7 \std{29.1} & 5.38 \std{9.16} & 0.02 \std{0.15} & 83.7 \std{30.6} & 2.33
  \std{6.04} & 5.40 \std{9.38} \\
  IM &
   43 & 89.4 & 80.6 \std{24.5} & 5.32 \std{12.70} & 0.21 \std{0.55} & 89.2 \std{25.2} & 1.28
  \std{3.98} & 5.44 \std{13.10} \\
  RCA &
   160 & 100.0 & 95.3 \std{5.0} & 2.18 \std{6.81} & 0.08 \std{0.33} & 98.2 \std{6.0} & 0.32
  \std{1.13} & 2.37 \std{7.28} \\
  R-PDA &
   150 & 98.8 & 82.6 \std{21.9} & 5.00 \std{9.10} & 0.12 \std{0.40} & 88.4 \std{22.3} & 1.35
  \std{4.52} & 5.15 \std{9.19} \\
  R-PLA &
   147 & 96.2 & 83.6 \std{18.6} & 5.92 \std{9.27} & 0.10 \std{0.39} & 89.4 \std{18.4} & 1.02
  \std{2.35} & 6.07 \std{9.43} \\
  L-PDA &
   8 & 100.0 & 75.1 \std{29.0} & 8.72 \std{18.10} & 0.25 \std{0.43} & 81.6 \std{31.8} & 2.82
  \std{6.76} & 8.99 \std{18.30} \\
  L-PLA &
   9 & 96.9 & 70.9 \std{27.2} & 8.32 \std{12.50} & 0.22 \std{0.42} & 77.7 \std{30.3} & 2.50
  \std{5.02} & 8.81 \std{13.00} \\
    Other &
   14 & 88.8 & 81.3 \std{17.0} & 10.14 \std{15.80} & 0.14 \std{0.35} & 86.7 \std{16.2} & 2.10
  \std{3.65} & 11.55 \std{17.00} \\
    \midrule
    All segments &
  160 & 95.3 & 92.8 \std{3.1} & 2.46 \std{3.62} & 0.2 \std{0.4} & 95.4 \std{3.6} & 0.53
  \std{0.33} & 4.58 \std{5.75} \\
    \bottomrule
    \end{tabular}
    \end{adjustbox}
    \caption{Inter-observer segmentation agreement per coronary segment. Agreement is the percentage of scans in which both analysts either annotate that segment or both omit it. Remaining metrics were computed over the $n$ scans in which both analysts annotated the segment. DSC: Dice similarity coefficient (\%); $\beta^{err}$: Betti number error; HD95: 95th percentile Hausdorff Distance (mm); ASSD: Average symmetric surface distance (mm). Coronary segment names can be found in Supplementary Table 1 and Figure 2.}
    \label{tab:inter_segment}
    \end{table*}
    
\subsubsection*{Label reliability across scan and vessel characteristics}

We analysed the effect of atherosclerotic plaque, scan quality, coronary dominance, coronary segment, lumen attenuation, diameter and distal position on inter-observer variability. 

\begin{figure}[h]
    \centering
    \includegraphics[width=1\linewidth]{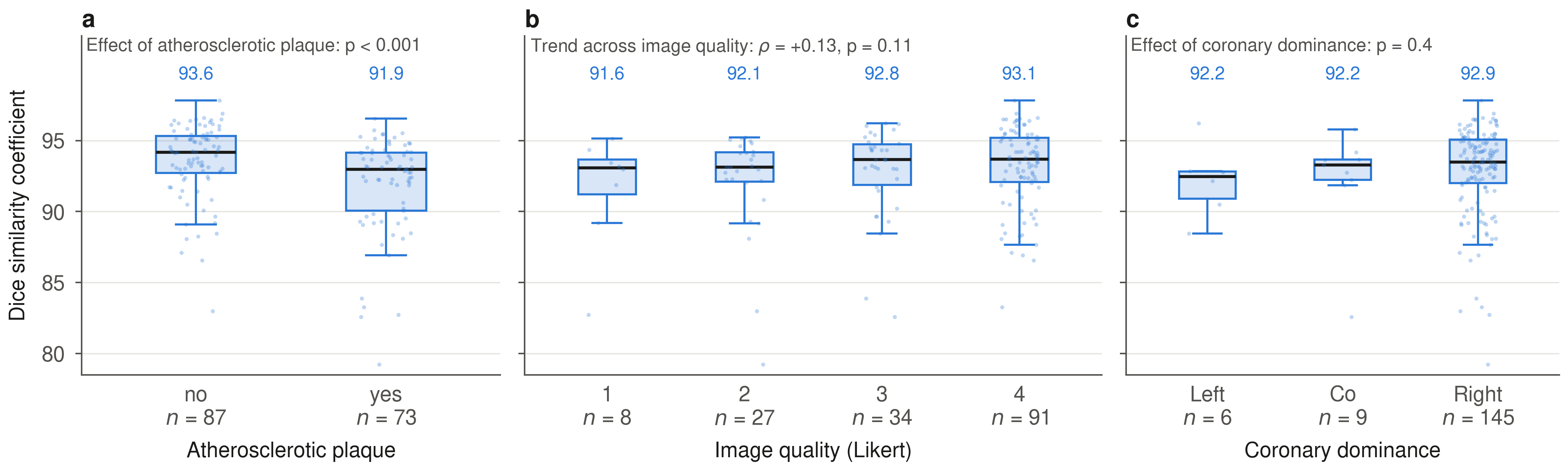}
    \caption{Effect of disease (a), image quality (b) and coronary dominance (c) on analyst segmentation performance. Each data point represents a single CCTA scan.}
    \label{fig:disease_quality_dom}
\end{figure}

A comparison of DSC performance stratified by the presence of atherosclerotic plaque is presented in Fig.~\ref{fig:disease_quality_dom}(a). Segmentation performance is significantly lower in patients with atherosclerotic plaque than without (91.9 vs. 93.6, $p < 0.001$). The reduction in segmentation performance is expected as plaque tissue components can have similar attenuation to contrast material and large atherosclerotic lesions result in more complex lumen morphology making lumen differentiation more challenging. To assess the impact of image artefacts on lumen segmentation, we stratify DSC performance by image quality using the previously generated Likert grades (1--4). The results presented in Fig.~\ref{fig:disease_quality_dom}(b) indicate a small positive correlation between DSC and image quality ($\rho = +0.13$, $p = 0.11$). The correlation is small and not significant reflecting that artefacts typically affect small regions of the scan and therefore only have a relatively small effect on the final segmentation performance. The effect of coronary dominance on segmentation performance is presented in Fig.~\ref{fig:disease_quality_dom}(c). There are no significant differences between left, right and co-dominance ($p = 0.40$), which is expected as all coronary dominance types can contain challenging segments. 

In Table~\ref{tab:inter_segment}, we compare how segmentation performance varies across coronary segment. There is 95.3\% agreement between analysts on which segments are present, and high agreement (96.9--100.0\%) for the challenging PDA and PLA segments. There are significant differences in segmentation performance between the coronary segment for all metrics ($p < 0.001$). Specifically, the main branches (LM, LAD, LCx and RCA) were the best segmented (DSC 84.8--95.3), while the side branches were consistently worse (70.9--83.6).
We reason that this is because they are smaller in diameter and their lumen attenuation is closer to that of the surrounding tissue, making their boundaries less clear. This ambiguity leaves room for interpretation as to how far distally a vessel should be traced and whether a given branch should be included at all, and these judgements can vary between analysts.

\begin{figure}[h]
     \centering
     \includegraphics[width=1\linewidth]{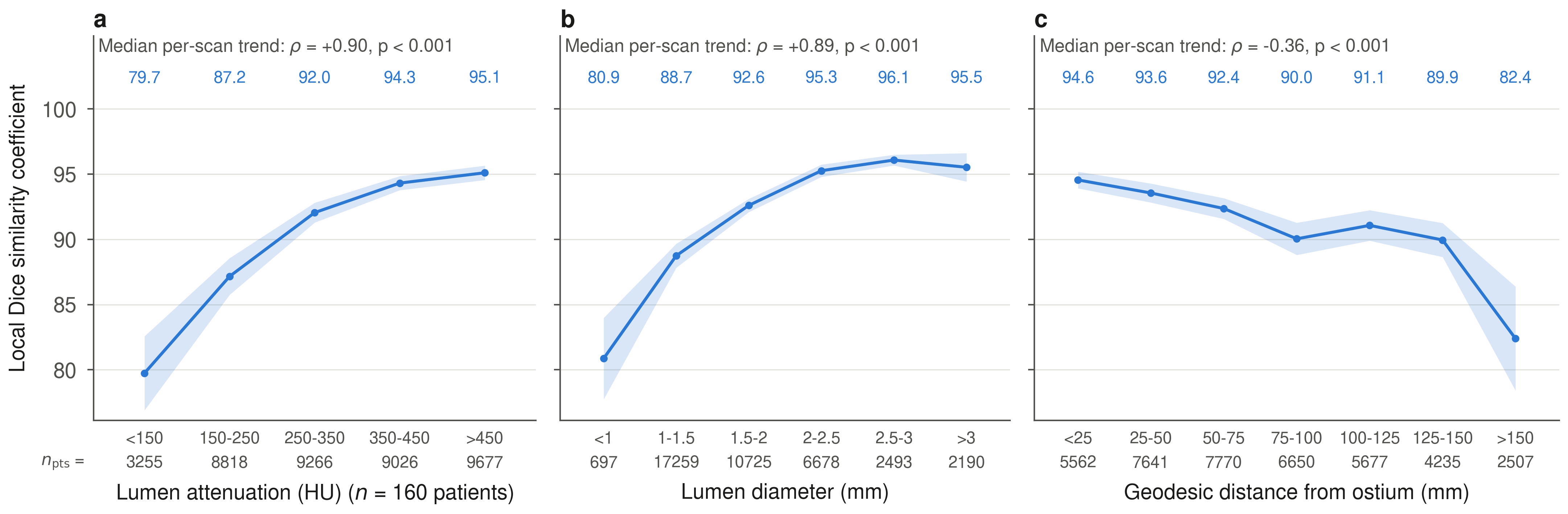}
     \caption{Effect of lumen attenuation (a), lumen diameter (b) and distal position (c) on analyst segmentation performance. Measured by local DSC on $8 \times 8 \times 8$ mm patch samples. }
     \label{fig:HU_diam_dist}
 \end{figure}

In addition, we analyse how local DSC performance varies with lumen attenuation, lumen diameter, and as we move from the aortic ostium distally along the coronary arteries. The results presented in Fig.~\ref{fig:HU_diam_dist} demonstrate that the local DSC score increases with lumen attenuation ($\rho$ = + 0.90, $p$ < 0.001), lumen diameter ($\rho$ = + 0.89, $p$ < 0.001), and decreases as we move distally along the vessel ($\rho$ = - 0.36, $p$ < 0.001). All three variables can be explained as the concentration of contrast material and the diameter of the vessel decrease distally making differentiation from fibrous, adipose, or muscular tissue more challenging.

 \begin{table*}[h]
\centering
\begin{adjustbox}{width=1\linewidth}
\begin{tabular}{lcccccc}
\toprule
&  \multicolumn{3}{c}{Lumen Segmentation} & \multicolumn{3}{c}{Lumen Centerline} \\
\cmidrule(lr){2-4} \cmidrule(lr){5-7}
 & DSC$\uparrow$ & HD95$\downarrow$ & $\beta^{err}$$\downarrow$ & clDice$\uparrow$ & ASSD$\downarrow$ & HD95$\downarrow$  \\
\midrule
TotalSegmentator~\cite{totalsegmentator} &  
70.5 \std{6.2} & 19.59 \std{6.45} & 4.6 \std{2.9} & 76.0 \std{5.3} & 
2.73 \std{0.90} & 23.26 \std{7.17} \\
3D-FFR-UNet~\cite{song2022automatic} &  
84.9 \std{5.5} & 15.93 \std{20.42} & 4.8 \std{3.5} & 89.8 \std{5.0} & 
1.71 \std{1.56} & 17.00 \std{18.23} \\
ADE-HTL~\cite{zhang2023anatomy} & 
87.7 \std{2.8} & \textbf{2.97 \std{3.74}} & \textbf{1.5 \std{1.5}} & \textbf{93.2 \std{3.2}} & 
0.74 \std{0.39} & \textbf{5.61 \std{4.85}} \\
Swin-UNETR~\cite{swinunetr} & 
87.9 \std{2.7} & 3.18 \std{3.68} & 3.8 \std{2.3} & 92.5 \std{3.0} & 
0.78 \std{0.36} & 6.16 \std{5.11} \\
ImageCAS~\cite{zeng_imagecas_2023} & 
87.9 \std{2.9} & 4.45 \std{4.90} & 4.7 \std{3.1} & 91.7 \std{3.5} & 0.90 \std{0.46} & 7.81 \std{6.03} \\
nnU-Net~\cite{isensee2021nnu} & 
89.8 \std{3.2} & 7.08 \std{12.65} & 5.6 \std{3.5} & 
92.3 \std{3.6} & 1.02 \std{0.75} & 10.41 \std{13.41} \\
nnU-Net + clDice~\cite{shit2021cldice} & 
90.0 \std{3.5} & 9.70 \std{15.36} & 8.0 \std{4.4} & 91.7 \std{3.9} & 1.20 \std{0.99} & 12.95 \std{14.20} \\
CAS-Net~\cite{dong2023novel} & 
\textbf{91.2 \std{2.8}} & 2.99 \std{3.47} & 1.9 \std{1.5} & 93.3 \std{3.2} & 
\textbf{0.73 \std{0.36}} & 5.75 \std{4.98} \\
\midrule
Inter-observer &  
92.8 \std{3.1} & 2.46 \std{3.62} & 0.4 \std{0.4} & 95.4 \std{3.6} & 0.53 \std{0.33} & 4.58 \std{5.75} \\
ImageCAS (labels)~\cite{zeng_imagecas_2023} &  
41.8 \std{6.7} & 16.15 \std{8.25} & 7.0 \std{6.7} & 78.2 \std{6.9} & 
2.23 \std{0.94} & 18.89 \std{8.91} \\
\bottomrule
\end{tabular}
\end{adjustbox}
\caption{Quantitative comparison of vessel segmentation algorithms (top) and expert analysts (bottom). Bold indicates the best performing algorithm. DSC: Dice similarity coefficient (\%); $\beta^{err}$: Betti number error; HD95: 95th percentile Hausdorff Distance (mm); ASSD: Average symmetric surface distance (mm).}
\label{tab:quant_main}
\end{table*}

\begin{figure}[h]
    \centering
    \includegraphics[width=1\linewidth]{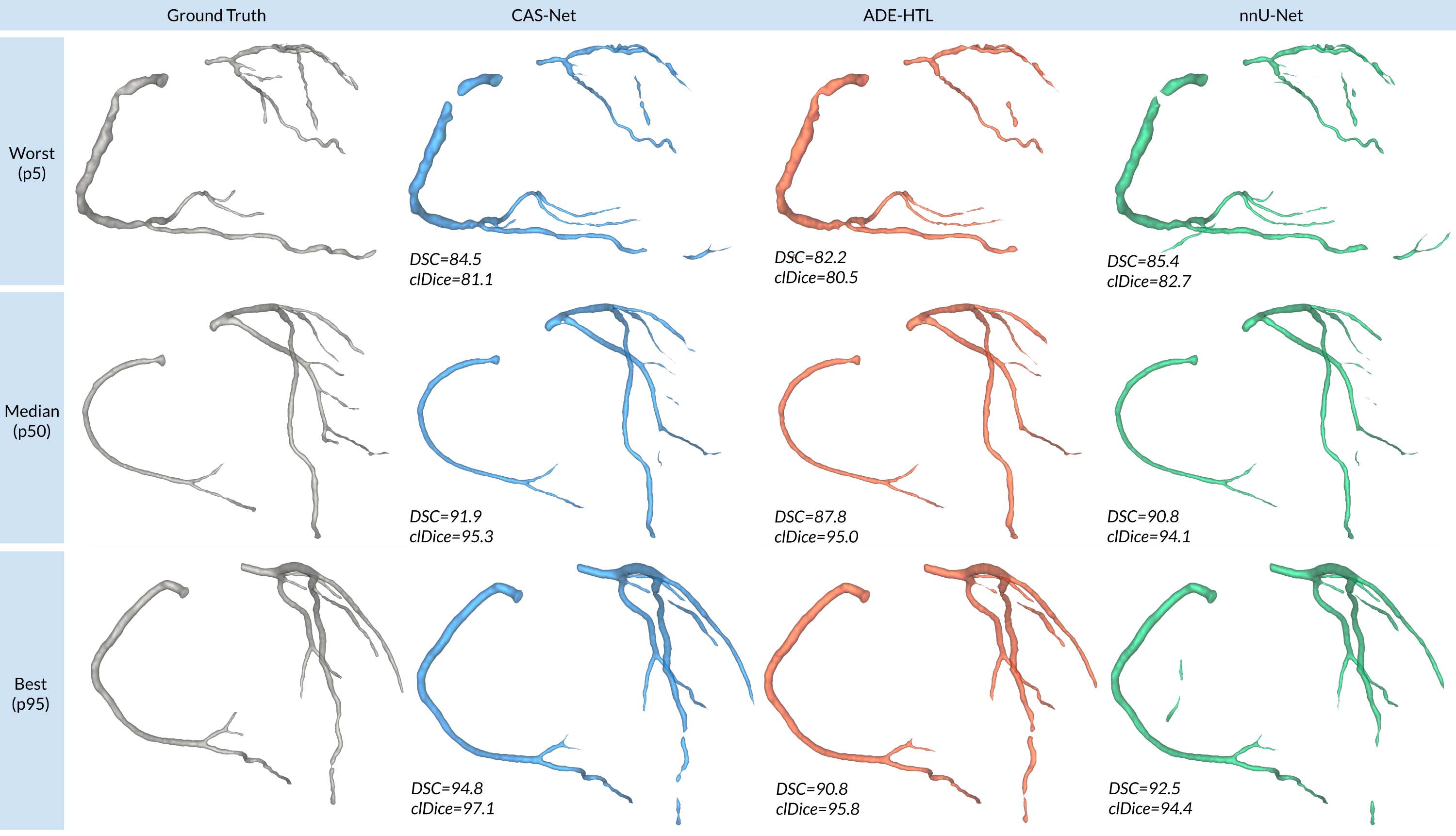}
    \caption{Qualitative comparison between model predictions using three examples that reflect worst, median and best model performance.}
    \label{fig:qual}
\end{figure}

\subsubsection*{Comparison between automated methods and human analysts}

In Table~\ref{tab:quant_main}, we present a quantitative comparison of all automated methods alongside inter-observer variability for lumen segmentation and centerline extraction. The best-performing automated method was CAS-Net with a DSC of 91.2, HD95 of 2.99 mm, $\beta^{err}$ of 1.9 for lumen segmentation, and a clDice of 93.3, ASSD of 0.73 mm and HD95 of 5.75 mm for centerline extraction. A statistically significant difference was found between the inter-observer variability and CAS-Net in DSC ($p < 0.001 $), HD95 ($p = 0.04$) and $\beta^{err}$ ($p < 0.001 $) for lumen segmentation, and clDice ($p < 0.001 $), ASSD ($p < 0.001$) and HD95 ($p = 0.001$) for centerline extraction, indicating that automated methods do not yet match human-level agreement on this dataset. We attribute this gap to three factors. First, analysts use surrounding context such as perivascular tissue, plaque and coronary veins to delineate the lumen, while the automated methods can implicitly encode this information they do not receive explicit supervision for these structures. Second, analysts followed a predefined protocol that excluded branches on the basis of the 18-segment model, position within the arterial tree and vessel diameter; these rules are not encoded in the training objective of the automated methods. Third, analysts corrected topological errors arising from image artefacts and ambiguous regions by reasoning about global tree connectivity, a mechanism the automated methods lack. Furthermore, both annotators edited the same automatically generated centerlines and initialised their segmentations using a 3D U-Net which predicted the same boundaries for unedited centerlines, and therefore the DSC inter-observer variability should be considered the upper-bound of agreement. 

A statistically significant difference was found between CAS-Net and the next best-performing method (ADE-HTL) for the DSC metric ($p < 0.001$) and $\beta^{err}$ ($p < 0.001$), but not for HD95 ($p = 0.87$) in the lumen segmentation task, and clDice metric ($p = 0.94$), ASSD ($p = 0.74$) and HD95 ($p = 0.90$) in the centerline extraction task. TotalSegmentator was the only pretrained model and performed the worst ($p < 0.001$, compared to CAS-Net), which could be attributed to the quality of the annotations it was trained on, the annotation protocol used, or a data distribution shift between its training data and ImageCAS. 

In Fig.~\ref{fig:qual}, we present three qualitative comparisons of lumen segmentation using examples that reflect the worst (5th percentile), median and best (95th percentile) model performance. These examples demonstrate that topological errors such as vessel breaks are present in all model predictions despite high DSC and clDice.

In Table~\ref{tab:effiency} we compare the computational efficiency of all methods and the analysts at inference time. CAS-Net is also the most computationally efficient. All models predicted the lumen segmentation in < 2 minutes which is significantly faster than the analysts who averaged 35 minutes per scan. 

\begin{table}[h]
\centering
\begin{adjustbox}{width=0.75\linewidth}
\begin{tabular}{lcccc}
 \toprule
 & \#Params (M) & Memory (GB) & \#Models & Inference (s) \\ 
\cmidrule{1-5}
TotalSegmentator~\cite{totalsegmentator} &
88.6 & 10.9 & 1 & 12.0  \\
3D-FFR-UNet~\cite{song2022automatic} &
17.7 & 2.7 & 2 & 38.6  \\
ADE-HTL~\cite{zhang2023anatomy} & 
10.7 & 2.7 & 1 & 41.6 \\
Swin-UNETR~\cite{swinunetr} &
70.2 & 8.5 & 1 & 23.7  \\
ImageCAS~\cite{zeng_imagecas_2023} &
23.3 & 25.9 & 3 & 14.9  \\
nnU-Net~\cite{isensee2021nnu} &
153.9 & 2.5 & 5 & 102.3  \\
nnU-Net + clDice~\cite{shit2021cldice} &
153.9 & 2.5 & 5 & 102.3  \\
CAS-Net~\cite{dong2023novel} &
6.8  & 9.1 & 1 & 15.5  \\
\midrule
Analysts &
-  & - & - & 2100.0  \\
\bottomrule
\end{tabular}
\end{adjustbox}
\caption{Comparison of inference time efficiency between methods. \#Params is the number of trainable parameters summed across all models, memory is the peak GPU memory, and inference time is measured from preprocessed volume to final prediction. M: Million, GB: Gigabytes, s: seconds.}
\label{tab:effiency}
\end{table}

\section*{Data Availability}

The dataset is available to download via Zenodo (\url{https://zenodo.org/records/21887809})~\cite{bransby_zenodo}. The dataset generated in this study is subject to a CC BY 4.0 license. The volumes from original ImageCAS are not re-distributed and are provided by the original authors via Kaggle (\url{https://www.kaggle.com/datasets/xiaoweixumedicalai/imagecas}) under the Apache 2.0 license. 

\section*{Code Availability}

Training and benchmarking code is available at \url{https://github.com/kitbransby/ImageCAS-X}. Pretrained weights are available via \url{https://zenodo.org/records/21887809}. 


\section*{Author Contributions}

Conceptualization: K.M.B, R.R.P, K.F.K; Methodology: K.M.B; Software: K.M.B, R.R.P; Validation: K.M.B; Formal analysis: K.M.B; Investigation: K.M.B; Data curation: K.M.B, E.Ø, K.K, J.K, A.J; Supervision: R.R.P, K.F.K, P.R.P, Y.E.Y, M.d.K; Project administration: R.R.P, K.F.K; Funding acquisition: R.R.P, K.F.K; Writing – original draft: K.M.B; Writing – review \& editing: all authors

\section*{Competing Interests}

K.F.K has received research grants from AP Møller og hustru Chastine McKinney Møllers Fond, Novo Nordisk Foundation, Novo Nordic A/S, Sygeforsikringen Danmark, Research Council of Rigshospitalet, The University of Copenhagen, Canon Medical Systems, GE Healthcare. Remaining authors declare no competing interests. 

\section*{Funding}

Funding for this research was provided by Novo Nordisk A/S. The funder had no role in study design, data collection, analysis, or preparation of the manuscript.

\clearpage
\appendix

\begin{center}
{\Large\bfseries ImageCAS-X - Supplementary Material}
\end{center}

\section{Segmentation accuracy in the ImageCAS dataset}

Three visual examples are presented in Supplementary Fig.~\ref{fig:imagecas_issues} to explain why the ImageCAS (2023) dataset is not suitable for training and validation of lumen segmentation methods in CCTA.

\begin{figure}[h]
    \centering
    \includegraphics[width=1\linewidth]{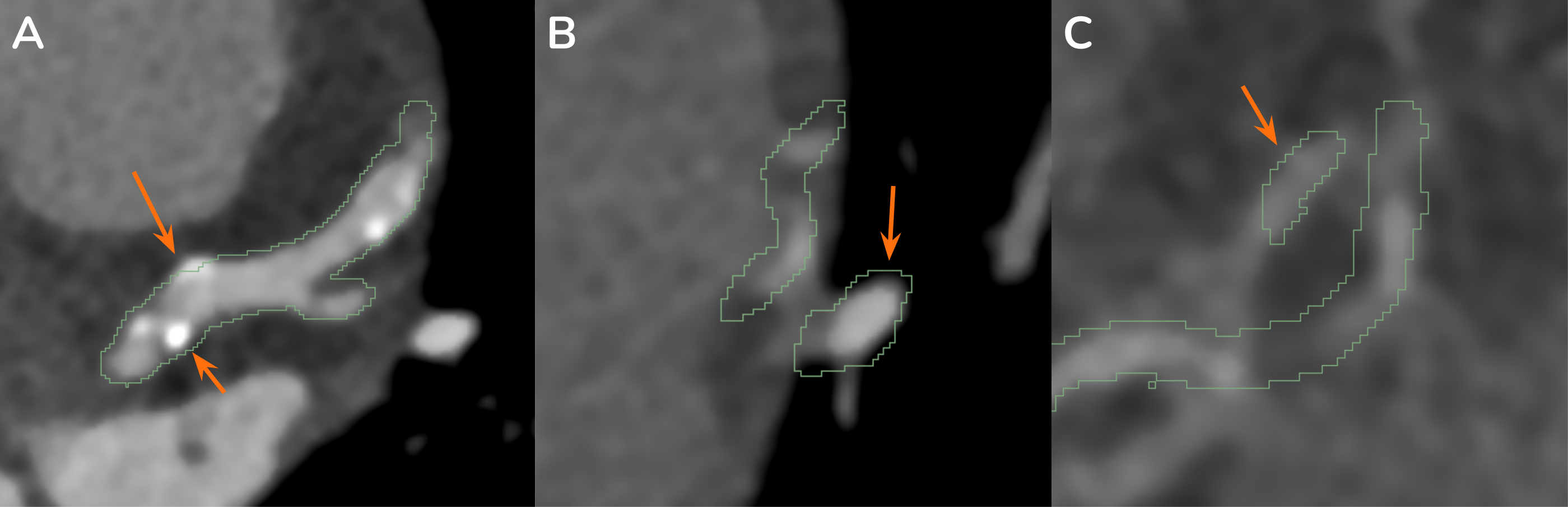}
    \caption{Segmentation issues on ImageCAS (2023) dataset (patient 4) such as the inclusion of atherosclerotic plaque (A), false positive pulmonary blood vessels (B), false positive coronary veins (C). }
    \label{fig:imagecas_issues}
\end{figure}

\section{Coronary segment labelling}

In this section we provide further detail on how the coronary trees were labelled and how ambiguous cases were resolved. 

Side branches were not traced where the connection to the parent vessel was unclear, where the branch could not be clearly delineated, or where the diameter at the most distal point was <1 mm. Septal perforators, acute marginals and nodal arteries were not traced. Additional diagonal (D) and obtuse marginal (OM) branches, such as D3, D4, OM3 and OM4, were labelled as `Other' if they were at least as large as the first or second branches or >1.8 mm in diameter at the most proximal end. Where a side branch itself bifurcated, only the larger branch was traced, unless the two branches were of equal size or the smaller was >1.8 mm in diameter, in which case both were traced. The same rule was applied where multiple posterior descending (PDA) or posterior lateral (PLA) arteries were present.

\begin{figure}[h]
    \centering
    \includegraphics[width=1\linewidth]{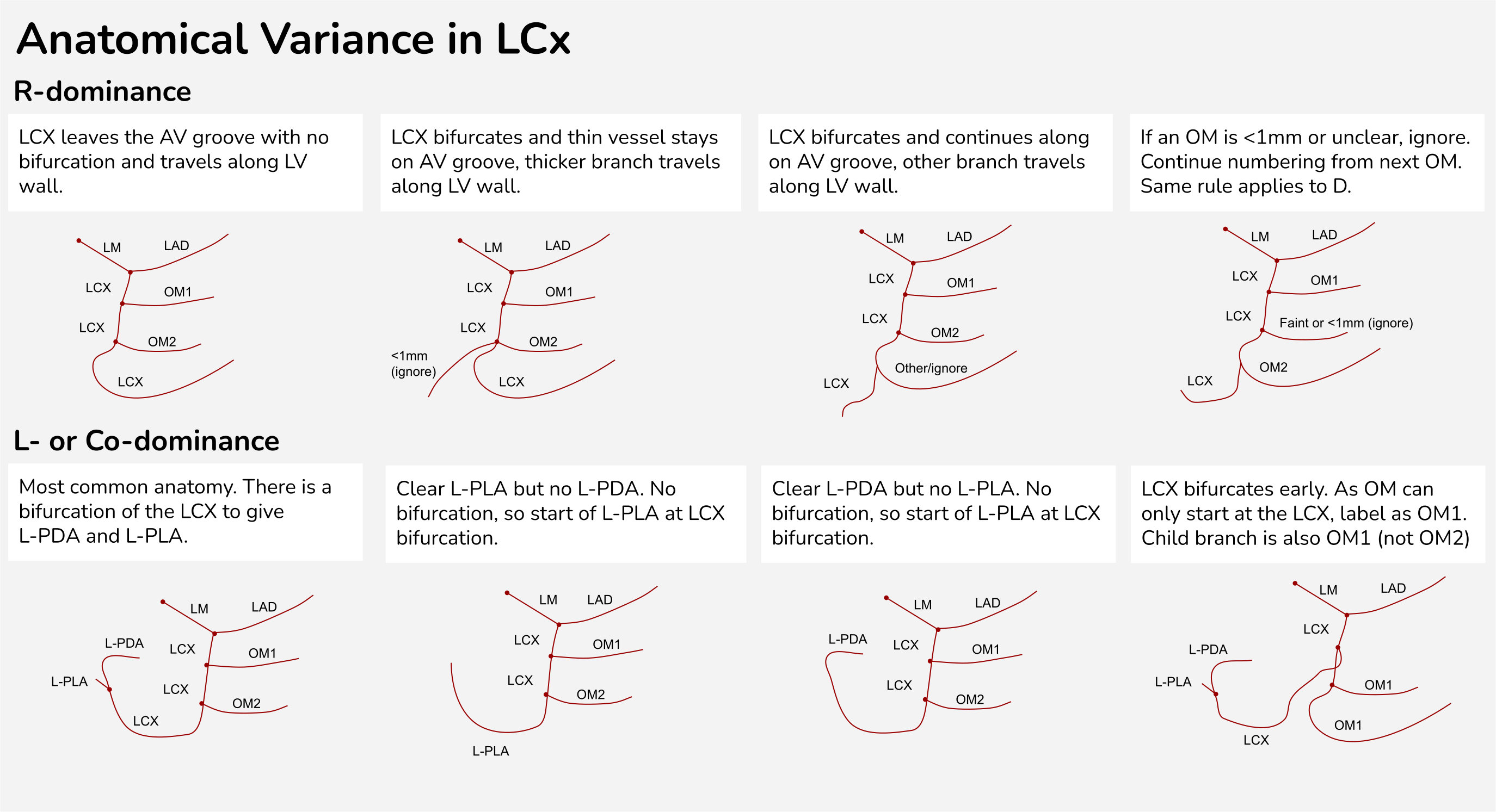}
    \caption{Labelling guide for anatomical variation in the left circumflex artery.}
    \label{fig:anatomy}
\end{figure}

Where a main branch bifurcated, a decision was required as to which branch was the side branch and which was the continuation of the main vessel. In all such cases this was determined by the direction and course of the artery rather than by its size. For example, where the LCx bifurcated, the branch travelling towards the left ventricular wall was labelled the OM and the branch continuing along the left atrioventricular groove was labelled the LCx, even where the OM was the larger of the two. The same argument was applied to the LAD where the branch that travelled along the anterior interventricular groove (AIVG) towards cardiac apex was labelled as the LAD, while the branch that travelled to left ventricular wall was labelled as the diagonal branch. Supplementary Fig.~\ref{fig:anatomy} illustrates the labelling decisions applied to anatomical variation in the LCx, the most challenging branch.

\section{Initial lumen generation}

Due to the significant labour involved in segmenting the coronary lumen from scratch, a 3D U-Net was trained to generate initial contour segmentations. Lumen contours were manually drawn in the cMPR volume from 80 randomly selected vessels from the train set, and a further 20 randomly selected vessels from the validation set were used to select the best model. The model was based on the nnU-Net architecture with five levels with feature dimensions: [32, 64, 128, 256, 512], resulting in 22.2M parameters. Random crops from cMPR volume of size $N \times H \times W$ were used as input, where $N$ was the number of cMPR frames set at 10 with 0.4mm spacing between frames, and $H$ and $W$ were the height and width of the plane, set at 64 representing a $16 \times 16 mm$ cross section using 0.25 mm pixel spacing. The model was trained using the same recipe and hardware described in the benchmark section of the main paper. At inference, a sliding window was used with step size of 1, and the probabilities from multiple windows were averaged to obtain a final prediction for each frame. 

\section{Coronary segment class indices}

Mapping between segment class index and segment name used in this study is presented in Supplementary Table~\ref{tab:segment_naming}.  

\begin{table}[h]
    \centering
    \begin{adjustbox}{width=0.55\linewidth}
    \begin{tabular}{llc}
    \toprule
        Branch & Segment & Class ID \\
    \midrule
        Left main & LM & 1 \\
        Left anterior descending artery & LAD & 2 \\
        Left circumflex artery & LCx & 3 \\
        First diagonal branch & D1 & 4 \\
        Second diagonal branch & D2 & 5 \\
        First obtuse marginal branch & OM1 & 6 \\
        Second obtuse marginal branch & OM2 & 7 \\
        Intermediate marginal (ramus intermedius) & IM & 8 \\
        Right coronary artery & RCA & 9 \\
        Right posterior descending artery & R-PDA & 10 \\
        Right posterior lateral artery & R-PLA & 11 \\
        Left posterior descending artery & L-PDA & 12 \\
        Left posterior lateral artery & L-PLA & 13 \\
        Other & D3, D4, OM3, or OM4 & 14 \\
    \bottomrule
    \end{tabular}
    \end{adjustbox}
    \caption{Coronary segment names mapped to segmentation indices}
    \label{tab:segment_naming}
\end{table}

\section{Shared training and inference framework}

All methods are trained and evaluated through a single shared framework that
fixes everything which must remain identical for a fair comparison (preprocessing, augmentation, optimiser, schedule, training budget, post-processing,
test-time augmentation and metrics), while retaining method-specific components: architecture, loss, input sampling strategy, batch size
and learning rate. All experiments use a fixed random seed. The shared components are described in the following sections. 

\subsection{Inputs \& Sampling}

Every scan is resampled to isotropic $0.5$ mm spacing, and intensities are clipped to a $[-200, 1000]$ HU window and linearly mapped to $[0, 1]$. The framework supports three input types: whole volume where the full preprocessed volume is passed to the network; random crop where a fixed-size patch is cropped at a random location from the whole volume; and center crop where patches are extracted at precomputed coordinates (e.g. points predicted by an earlier stage) rather than at random locations. For all patch-based methods that take a random patch from each image, foreground
(fg) sampling is used to ensure at least $50\%$ of every batch contains patches with
lumen class. Without this, small patches ($64^3$--$96^3$) land on vessel so rarely under uniform cropping that batches are almost entirely background resulting in noisy gradients during backprop. 

\subsection{Augmentation}

Augmentation is applied during training and is based on nnU-Net's default
pipeline: additive Gaussian noise
($p{=}0.1$, $\sigma^2 \in [0, 0.01]$), Gaussian blur ($p{=}0.2$,
$\sigma \in [0.5, 1.0]$), multiplicative brightness ($p{=}0.15$, $[0.75, 1.25]$),
contrast ($p{=}0.15$, $[0.75, 1.25]$), simulated low resolution ($p{=}0.25$,
downsample factor $[0.5, 1.0]$), inverted gamma ($p{=}0.1$, $\gamma \in [0.7, 1.5]$)
and gamma ($p{=}0.3$, $\gamma \in [0.7, 1.5]$). Axis flipping over coronal and sagittal planes are included, but rotations and axial flipping are omitted since coronary anatomy has a genuine superior--inferior orientation that the network should not be trained to treat as ambiguous.

\subsection{Training schedule}

An epoch is defined as 250 training and 50 validation batches drawn with replacement. All methods are trained for the same number of epochs with Adam (weight decay $10^{-8}$) and a polynomial learning-rate decay (power $0.9$), apart from nnU-Net which used SGD. Learning rate
and batch size are the only optimisation hyperparameters set per method, so that
each can be tuned to its own architecture and memory footprint. Checkpoints are selected on validation performance, by default the mean lumen Dice
over validation batches. 

\subsection{Inference}

For inference, whole-volume methods used a single forward pass, while patch- and crop-based methods are tiled with a sliding window of the method's own training patch size at $50\%$
overlap where overlapping predictions are combined by a Gaussian-weighted average with
per-axis $\sigma = 0.125 \times$ patch size. Inference for centre-driven patch methods occurs only at precomputed centres. Axis mirroring test-time augmentation is applied uniformly to every method: predictions
are averaged over all four flip combinations of the two in-plane axes, matching the
flip axes used during training. Predicted logits are converted to probabilities and resampled back onto each scan's original acquisition grid by linear interpolation before discretisation. Post-processing is then applied at full original resolution and is identical for all methods: threshold at $0.5$, followed by removal of connected components smaller than 100 voxels. Component removal is size-based rather than keep-largest-$N$, as the LCx and LAD can connect directly to the aortic annulus resulting in variable number of true components. 

\section{Method-specific implementation notes}

In Supplementary Table~\ref{tab:hyperparams}, we summarise the final hyperparameters chosen for each baseline method. Further implementation details for each method can be found below. 

\begin{table}[h]
    \centering
    \begin{adjustbox}{width=0.8\linewidth}
    \begin{tabular}{llcccc}
    \toprule
        Method & Stage &  Batch Size & Learning Rate & Loss & \\
        \midrule
        3D-FFR-UNet & - & 4 & 0.0001 & Dice & \\
        ADE-HTL & ADE & 2 & 0.0001 & Dice, CE   & \\
           & HTL & 4 & 0.0001 & Dice, CE, Weighted HD, MSE& \\
        Swin-UNETR & - & 4 & 0.0001 & Dice & \\
        ImageCAS & Coarse & 8 & 0.0001 & Dice & \\
         & Dilated & 8 & 0.0001 & Weighted Similarity & \\
         & 16³ & 512 & 0.0001 &  Dice [DS] & \\
          & 32³ & 64 & 0.0001 &  Dice [DS] &\\
         & 64³ & 8 & 0.0001 &  Dice [DS] & \\
        nnU-Net & - & 2 & 0.01 & Dice, CE [DS]  & \\
        nnU-Net + clDice & - & 2 & 0.01 & Dice, CE, clDice [DS] & \\
        CAS-Net & - & 4 & 0.0001 & Dice  & \\
    \bottomrule
    \end{tabular}
    \end{adjustbox}
    \caption{Hyperparameter settings for baseline methods. CE: cross-entropy; DS: deep supervision; HD: Hausdorff distance; MSE: mean squared error.}
    \label{tab:hyperparams}
\end{table}

\subsection{3D-FFR-UNet} The authors propose two steps to first automatically identify axial slices containing coronary arteries which is used to crop the volume which is then patched to 64³ and fed into a second model. While this first step is reasonable, it is not necessary if foreground over-sampling is applied during training. As this is implemented in our patch data loader already, we skip this step. The authors experiment with many loss functions, and we follow their recommendations to use dice loss. 


\subsection{CAS-Net} We find suboptimal results when using their proposed loss recipe on our data and instead we train with a pure dice loss which gives good results. 

\subsection{ADE-HTL} Training their method on our hardware is slow and limited to a batch size of 1. Therefore we use key-value average pooling before their BAI module which shrinks the key set by $64\times$, and train with mixed precision. The authors use a weighted cross-entropy loss, but the weighting parameters are not given. As using the wrong weighting parameters is likely to result in wildly different predictions, we simplify by using dice and cross-entropy loss for the connectivity prediction. As dice is robust to class-imbalance, this is a reasonable solution. 

\subsection{ImageCAS (2023)} When ensembling final models, we ensemble the probabilities rather than majority voting on the binary segmentation masks as probabilities are richer representations. We find that excluding the stage 1 coarse segmentation model from the ensemble improves performance.


\begin{thebibliography}{99}

\bibitem{leipsic_scct_2014}
Leipsic, J., Abbara, S., Achenbach, S., Cury, R., Earls, J.P. et al. SCCT guidelines for the interpretation and reporting of coronary CT angiography: A report of the Society of Cardiovascular Computed Tomography Guidelines Committee. \textit{Journal of Cardiovascular Computed Tomography} \textbf{8}, 342–358 (2014).

\bibitem{nieman_standards_2024}
Nieman, K., García-García, H.M., Hideo-Kajita, A., Collet, C., Dey, D. et al. Standards for quantitative assessments by coronary computed tomography angiography (CCTA). \textit{Journal of Cardiovascular Computed Tomography} \textbf{18}, 429–443 (2024).

\bibitem{knuuti20202019}
Knuuti, J., Wijns, W., Saraste, A., Capodanno, D., Barbato, E. et al. 2019 ESC Guidelines for the diagnosis and management of chronic coronary syndromes: The Task Force for the diagnosis and management of chronic coronary syndromes of the European Society of Cardiology (ESC). \textit{European heart journal} \textbf{41}, 407–477 (2020).

\bibitem{writing20212021}
Writing Committee Members, Gulati, M., Levy, P.D., Mukherjee, D., Amsterdam, E. et al. 2021 AHA/ACC/ASE/CHEST/SAEM/SCCT/SCMR guideline for the evaluation and diagnosis of chest pain: a report of the American College of Cardiology/American Heart Association Joint Committee on Clinical Practice Guidelines. \textit{Journal of the American College of Cardiology} \textbf{78}, e187–e285 (2021).

\bibitem{williams_coronary_2019}
Williams, M.C., Moss, A.J., Dweck, M., Adamson, P.D., Alam, S. et al. Coronary Artery Plaque Characteristics Associated With Adverse Outcomes in the SCOT-HEART Study. \textit{Journal of the American College of Cardiology} \textbf{73}, 291–301 (2019).

\bibitem{scotheart}
Williams, M.C., Kwiecinski, J., Doris, M., McElhinney, P., D’Souza, et al. Low-attenuation noncalcified plaque on coronary computed tomography angiography predicts myocardial infarction: results from the multicenter SCOT-HEART trial. \textit{Circulation} 141(18) pp.1452-1462 (2020)

\bibitem{thomsen2016characteristics}
Thomsen, C. \& Abdulla, J. Characteristics of high-risk coronary plaques identified by computed tomographic angiography and associated prognosis: a systematic review and meta-analysis. \textit{European Heart Journal-Cardiovascular Imaging} \textbf{17}, 120–129 (2016).

\bibitem{oikonomou_non-invasive_2018}
Oikonomou, E.K., Marwan, M., Desai, M.Y., Mancio, J., Alashi, A. et al. Non-invasive detection of coronary inflammation using computed tomography and prediction of residual cardiovascular risk (the CRISP CT study): a post-hoc analysis of prospective outcome data. \textit{The Lancet} \textbf{392}, 929–939 (2018).

\bibitem{antonopoulos_detecting_2017}
Antonopoulos, A.S., Sanna, F., Sabharwal, N., Thomas, S., Oikonomou, E.K. et al. Detecting human coronary inflammation by imaging perivascular fat. \textit{Science Translational Medicine} \textbf{9}, eaal2658 (2017).

\bibitem{norgaard_diagnostic_2014}
Nørgaard, B.L., Leipsic, J., Gaur, S., Seneviratne, S., Ko, B.S. et al. Diagnostic Performance of Noninvasive Fractional Flow Reserve Derived From Coronary Computed Tomography Angiography in Suspected Coronary Artery Disease. \textit{Journal of the American College of Cardiology} \textbf{63}, 1145–1155 (2014).

\bibitem{dong2023novel}
Dong, C., Xu, S., Dai, D., Zhang, Y., Zhang, C. et al. A novel multi-attention, multi-scale 3D deep network for coronary artery segmentation. \textit{Medical Image Analysis} \textbf{85}, 102745 (2023).

\bibitem{song2022automatic}
Song, A., Xu, L., Wang, L., Wang, B., Yang, X. et al. Automatic coronary artery segmentation of CCTA images with an efficient feature-fusion-and-rectification 3D-UNet. \textit{IEEE journal of biomedical and health informatics} \textbf{26}, 4044–4055 (2022).

\bibitem{qiu2025topology}
Qiu, Y., Shan, D., Wang, Y., Dong, P., Wu, D. et al. A topology-preserving three-stage framework for fully-connected coronary artery extraction. \textit{Medical Image Analysis} \textbf{103}, 103578 (2025).

\bibitem{yang2024segmentation}
Yang, X., Xu, L., Yu, S., Xia, Q., Li, H. et al. Segmentation and vascular vectorization for coronary artery by geometry-based cascaded neural network. \textit{IEEE Transactions on Medical Imaging} \textbf{44}, 259–269 (2024).

\bibitem{acc}
Chandrashekhar, Y., Blankstein, R., Shaw, L.J., Ferencik, M., Leipsic, J., et al. Quantitative Coronary Plaque Analysis Symposium Collaborators, 2025. Quantitative coronary plaque analysis in clinical practice: 2025 ACC scientific statement: a report of the American College of Cardiology. \textit{JACC: Cardiovascular Imaging} (2025)

\bibitem{shit2021cldice}
Shit, S., Paetzold, J.C., Sekuboyina, A., Ezhov, I., Unger, A. et al. clDice-a novel topology-preserving loss function for tubular structure segmentation. In \textit{Proceedings of the IEEE/CVF conference on computer vision and pattern recognition} 16560–16569 (2021).

\bibitem{metz20083d}
Metz, C., Schaap, M., van Walsum, T., van der Giessen, A., Weustink, A. et al. 3D segmentation in the clinic: A grand challenge II-coronary artery tracking. \textit{Insight Journal} \textbf{1}, 6 (2008).

\bibitem{kiricsli2013standardized}
Kiri{\c{s}}li, H., Schaap, M., Metz, C., Dharampal, A., Meijboom, W.B. et al. Standardized evaluation framework for evaluating coronary artery stenosis detection, stenosis quantification and lumen segmentation algorithms in computed tomography angiography. \textit{Medical image analysis} \textbf{17}, 859–876 (2013).

\bibitem{gharleghi_annotated_2023}
Gharleghi, R., Adikari, D., Ellenberger, K., Webster, M., Ellis, C. et al. Annotated computed tomography coronary angiogram images and associated data of normal and diseased arteries. \textit{Scientific Data} \textbf{10}, 128 (2023).

\bibitem{tu_mask_2025}
Tu, R., Tian, C., Wang, L., Deng, Y., Chen, C. et al. Mask SAM 3D for coronary artery and plaque segmentation in CCTA images. \textit{International Journal of Computer Assisted Radiology and Surgery} (2025).

\bibitem{zeng_imagecas_2023}
Zeng, A., Wu, C., Lin, G., Xie, W., Hong, J. et al. ImageCAS: A large-scale dataset and benchmark for coronary artery segmentation based on computed tomography angiography images. \textit{Computerized Medical Imaging and Graphics} \textbf{109}, 102287 (2023).

\bibitem{clough2020}
Clough, J.R., Byrne N., Oksuz I., Zimmer V.A., Schnabel J.A., et al. A topological loss function for deep-learning based image segmentation using persistent homology. \textit{IEEE transactions on pattern analysis and machine intelligence}. 8766-8778 (2020).

\bibitem{bransby2023}
Bransby, K.M., Tufaro, V., Cap, M., Slabaugh, G., Bourantas, C. et al. 3D coronary vessel reconstruction from bi-plane angiography using graph convolutional networks. \textit{IEEE 20th International Symposium on Biomedical Imaging (ISBI)} (2023)

\bibitem{wang2025deepca}
Wang, Y., Banerjee, A., Choudhury, R.P. and Grau, V., DeepCA: Deep Learning-Based 3D Coronary Artery Tree Reconstruction from Two 2D Non-Simultaneous X-Ray Angiography Projections. In \textit{2025 IEEE/CVF Winter Conference on Applications of Computer Vision (WACV}) 337-346 (2025)

\bibitem{feldman2025}
Feldman, P., Sinnona, M., Delrieux, C., Siless, V. and Iarussi, E., VesselGPT: Autoregressive Modeling of Vascular Geometry. In \textit{International Conference on Medical Image Computing and Computer-Assisted Intervention} 662-672 (2025)

\bibitem{baldachowski_coronary_3d_ct}
Baldachowski, M. \& Korona, M. Tracing the Coronary Arteries in 3D Computed Tomography Scans. Master's thesis, Technical University of Denmark (DTU Compute) (2024).

\bibitem{cpr}
Kanitsar, A., Fleischmann, D., Wegenkittl, R., Felkel, P. and Groller, E., CPR-curved planar reformation. \textit{IEEE Conference on Visualization} 37-44 (2002)

\bibitem{skeletonisation}
Lee, T. C., Kashyap, R. L., \& Chu, C. N. Building skeleton models via 3-D medial surface axis thinning algorithms. CVGIP: graphical models and image processing, 56(6), 462-478. (1994).

\bibitem{totalsegmentator}
Wasserthal, J., Breit, H.C., Meyer, M.T., Pradella, M., Hinck, et al. TotalSegmentator: robust segmentation of 104 anatomic structures in CT images. \textit{Radiology: Artificial Intelligence} 5(5), 2023.

\bibitem{bransby_zenodo}
Bransby, K. M. \& Paulsen, R. R. ImageCAS-X. Zenodo https://doi.org/10.5281/zenodo.21887809 (2026)

\bibitem{zhang2023anatomy}
Zhang, X., Sun, K., Wu, D., Xiong, X., Liu, J. et al. An anatomy-and topology-preserving framework for coronary artery segmentation. \textit{IEEE Transactions on Medical imaging} \textbf{43}, 723–733 (2023).

\bibitem{isensee2021nnu}
Isensee, F., Jaeger, P.F., Kohl, S.A., Petersen, J. \& Maier-Hein, K.H. nnU-Net: a self-configuring method for deep learning-based biomedical image segmentation. \textit{Nature methods} \textbf{18}, 203–211 (2021).

\bibitem{swinunetr}
Hatamizadeh, A., Nath, V., Tang, Y., Yang, D., Roth, H.R. et al. Swin unetr: Swin transformers for semantic segmentation of brain tumors in mri images. \textit{In International MICCAI brain lesion workshop} 272-284 (2021).

\bibitem{deformCL}
Zhao, Z., Zhang, Z., Liu, Y., Zhang, Z., Yu, H., et al. DeformCL: Learning Deformable Centerline Representation for Vessel Extraction in 3D Medical Image. \textit{In Proceedings of the Computer Vision and Pattern Recognition Conference} 30896-30905 (2025).

\bibitem{taha2015}
Taha, A.A. and Hanbury, A., Metrics for evaluating 3D medical image segmentation: analysis, selection, and tool. \textit{BMC medical imaging} \textbf{15(1)}, 29 (2015)

\bibitem{wasserman2018topological}
Wasserman, L. Topological data analysis. \textit{Annual review of statistics and its application} \textbf{5}, 501–532 (2018).
 
\end{thebibliography}
\end{document}